\documentclass{article}
\usepackage{amssymb}

\usepackage[preprint]{corl_2025} % Uncomment for pre-prints (e.g., arxiv); This is like ``final'', but will remove the CORL footnote.
\usepackage{moreverb,url}
\usepackage{xspace}
\usepackage{xcolor}
\usepackage{graphicx}
\usepackage{enumitem}

\usepackage{amssymb}
\usepackage{amsmath}
\usepackage{amsthm}
\usepackage{amsbsy}
\usepackage{bbm}

\usepackage{graphicx}
\usepackage{wrapfig}
\usepackage{booktabs} % for professional tables
\usepackage{multirow}
\usepackage{colortbl}

\usepackage{subfiles}
\usepackage{csquotes}
\usepackage{makecell}

\usepackage{graphicx}
\usepackage{caption}
\usepackage{subcaption}

\usepackage{array}

\newcommand\BibTeX{{\rmfamily B\kern-.05em \textsc{i\kern-.025em b}\kern-.08em
T\kern-.1667em\lower.7ex\hbox{E}\kern-.125emX}}

\definecolor{wine}{RGB}{204, 0, 102}
\definecolor{ocean}{RGB}{13, 121, 202}
\definecolor{light_ocean}{RGB}{18, 178, 235}
\definecolor{dark_ocean}{RGB}{10, 89, 148}
\definecolor{grey}{RGB}{170, 170, 170}
\definecolor{light-grey}{RGB}{220, 220, 220}
\definecolor{dark_gray}{rgb}{0.2, 0.2, 0.2} 
\definecolor{med-grey}{rgb}{0.3, 0.3, 0.3} 
\definecolor{grape}{RGB}{112,48,160}
\definecolor{aqua}{RGB}{52,172,139}
\definecolor{dark_aqua}{RGB}{35,115,93}
\definecolor{dark_orange}{RGB}{216,92,0}
\definecolor{vibrant_orange}{RGB}{255, 102, 0}
\definecolor{vibrant_blue}{RGB}{14, 120, 255}
\definecolor{vibrant_pink}{RGB}{255, 0, 104}
\definecolor{dark_red}{RGB}{122, 0, 0}
\definecolor{dark_green}{RGB}{0, 92, 34}

\definecolor{lightgreen}{RGB}{220, 255, 220}
\definecolor{lightorange}{RGB}{255, 240, 220}
\definecolor{lightblue}{RGB}{220, 240, 255}

\newcommand{\para}[1]{\medskip\noindent\textbf{#1. }}

\newcommand{\ours}{\textcolor{orange}{\textbf{ABA}}\xspace}
\newcommand{\vanilla}{\textcolor{dark_gray}{\textbf{Vanilla}}\xspace}
\newcommand{\policyem}{\textcolor{dark_ocean}{\textbf{PolicyEmbed}}\xspace}
\newcommand{\dino}{\textcolor{wine}{\textbf{DINOEmbed}}\xspace}

\newcommand{\figref}{Fig.~\ref}
\newcommand{\secref}{Sec.~\ref}

\newcommand{\funccor}{f}

\newcommand{\action}{a}
\newcommand{\actiontraj}{\mathbf{a}}
\newcommand{\actionSpace}{\mathcal{A}}

\newcommand{\latent}{z}
\newcommand{\latentSpace}{\mathcal{Z}}
\newcommand{\latenttraj}{\mathbf{z}}
\newcommand{\enc}{\mathcal{E}}

\newcommand{\obs}{o}
\newcommand{\obstraj}{\mathbf{o}}
\newcommand{\obsSpace}{\mathcal{O}}

\newcommand{\env}{E}
\newcommand{\envSpace}{\mathbb{E}}

\newcommand{\stateSpace}{\mathcal{Q}}

\newcommand{\IDEnv}{E_{\text{ID}}}
\newcommand{\OODEnv}{E_{\text{OOD}}}
\newcommand{\IDEnvLabel}[1]{E_{\text{ID-#1}}}
\newcommand{\OODEnvLabel}[1]{E_{\text{OOD-#1}}}
\newcommand{\IDDataset}{\mathcal{D}_{\text{ID}}}

\newcommand{\IDObsSet}{\mathcal{O}_{\text{ID}}}
\newcommand{\OODObsSet}{\mathcal{O}_{\text{OOD}}}
\newcommand{\obsood}{\hat{\obs}}

\newcommand{\FuncCorrIDObsSet}{\mathcal{O}_{f}}

\title{Adapting by Analogy: OOD Generalization of Visuomotor Policies via Functional Correspondence}

\author{
  Pranay Gupta \quad\quad        Henny Admoni  \quad\quad       Andrea Bajcsy\\
  The Robotics Institute, \\
  Carnegie Mellon University, United States\\
Correspondence: \texttt{pranaygu@andrew.cmu.edu} \\
}

\begin{document}
\maketitle

%===============================================================================

\begin{abstract}
End-to-end visuomotor policies trained using behavior cloning have shown a remarkable ability to generate complex, multi-modal low-level robot behaviors. However, at deployment time, these policies still struggle to act reliably when faced with out-of-distribution (OOD) visuals induced by objects, backgrounds, or environment changes. Prior works in interactive imitation learning solicit corrective expert demonstrations under the OOD conditions---but this can be costly and inefficient. We observe that task success under OOD conditions does not always warrant novel robot behaviors. In-distribution (ID) behaviors can directly be transferred to OOD conditions that share functional similarities with ID conditions. For example, behaviors trained to interact with in-distribution (ID) pens can apply to interacting with a visually-OOD pencil. The key challenge lies in disambiguating which ID observations functionally correspond to the OOD observation for the task at hand. We propose that an expert can provide this OOD-to-ID functional correspondence. Thus, instead of collecting new demonstrations and re-training at every OOD encounter, our method: (1) detects the need for feedback by first checking if current observations are OOD and then identifying whether the most similar training observations show divergent behaviors, (2) solicits functional correspondence feedback to disambiguate between those behaviors, and (3) intervenes on the OOD observations with the functionally corresponding ID observations to perform deployment-time generalization. We validate our method across diverse real-world robotic manipulation tasks with a Franka Panda robotic manipulator. Our results show that test-time functional correspondences can improve the generalization of a vision-based diffusion policy to OOD objects and environment conditions with low feedback.

\end{abstract}

% Two or three meaningful keywords should be added here
\keywords{visuomotor policy, OOD generalization, test-time adaptation} 

%===============================================================================

% \input{sections/intro}

\section{Introduction}
\label{sec:intro}

\begin{figure}[t!]
    \centering
    \includegraphics[width = \textwidth]{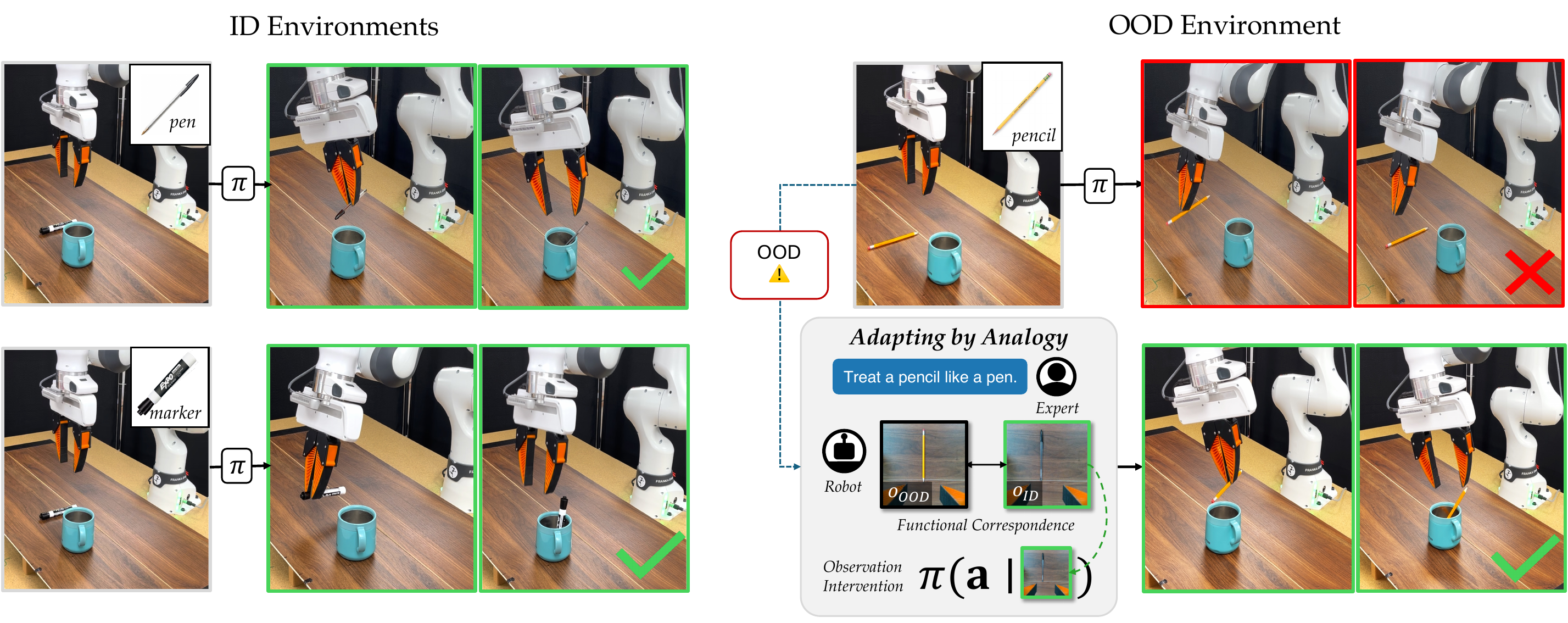}
    \caption{We present \textit{\textbf{Adapting by Analogy}}, a test-time method that uses functional correspondences between deployment and training conditions to improve a policy's performance in OOD conditions.}
    \label{fig:front-fig}
    \vspace{-1em}
\end{figure}

A central goal in robot learning is to enable robots to generalize: to successfully perform tasks in environments they have never seen before. 
Imagine a robot encountering a pencil for the first time. With just an RGB image, it should be able to reason about the scene and delicately place the object into a nearby cup, as shown in Figure~\ref{fig:front-fig}. 
One popular approach towards this is imitation-based visuomotor policy learning. 
However, while internet-scale datasets have powered generalization breakthroughs in vision and language, robotics still lacks access to the same data scale~\citep{khazatsky2024droid,  rt22023arxiv,contributors2024agibotworldrepo, fang2023rh20t, open_x_embodiment_rt_x_2023}, and collecting expert demonstration data remains expensive and time-consuming. 
This results in robots failing in unintuitive ways when faced with out-of-distribution (OOD) environments (lower left, Figure~\ref{fig:front-fig}). 
Nevertheless, even with modest expert demonstration datasets, recent advances in policy architectures and training algorithms have enabled robots to learn complex visuomotor skills---such as grasping thin tools or folding clothes and operating articulated objects---that work well in-distribution (ID)~\citep{chi2024diffusionpolicy,zhao2023learningfinegrainedbimanualmanipulation, kim24openvla,leebehavior,reuss2024multimodal}.
This raises the central question of our work: 
\textit{How can we reuse robot behaviors learned in in-distribution settings to succeed in out-of-distribution scenarios?}

Our key insight is that behavior generalization may not always require more demonstration data: it may just need a better correspondence between the training and test conditions. 
For example, in Figure~\ref{fig:front-fig}, even though the robot has never seen pencils before, it has seen similarly-thin pens and thicker markers (top row). Thus, if it understood that the pencil is functionally equivalent to the pen in this task, it could ``imagine'' that the pencil is a pen and reuse the pen pickup behavior to successfully complete the task. 
Based on this insight, we present \textit{\textbf{Adapting by Analogy (ABA)}}: a method which establishes \textit{functional correspondences} between in-distribution and out-of-distribution scenes to steer a visuomotor policy through OOD conditions. 
A key aspect of our method is to leverage expert human knowledge---in the form of a textual description---to interactively learn high-level functional correspondences relevant to the task at hand. 
The textual description is decoded into a functional correspondence feature space that matches corresponding semantic segments of the scene to retrieve ID behaviors that are "relevant" for the current OOD scene. 
To measure whether the functional correspondence is well-specified, the robot estimates its uncertainty over the retrieved behavior modes and continues to ask for correspondence refinement until it is certain in the mapping. 
% can be a powerful bridge for such knowledge transfer. 

% * Describe key results
We instantiate \textit{\textbf{Adapting by Analogy}} on hardware with a Franka Research 3 manipulator acting with a diffusion-based visuomotor policy \cite{chi2024diffusionpolicy}. 
By controlling the training and test environments, we study i) how functional correspondences can improve the task success rate in increasingly OOD environments, ii) if our method seeks expert feedback efficiently, and iii) we verify how critical functional correspondences are for OOD generalization. We find that even a relatively small number of expert-guided functional correspondences can significantly improve the generalization capabilities of a visuomotor policy interacting with OOD objects from new semantic categories. 
% We evaluate our method on two real-world tasks of varying complexity, where we introduce OODness through novel objects and environments. Our method upholds task success rate better than the baselines, while being efficient at seeking expert feedback.

% * Summarize your contribution
% In summary, in this work we test how far we can take the idea of leveraging functional correspondences for zero-shot behavior transfer.
% We present a novel method that leverages functional correspondences to transfer behaviors from ID to OOD. It also serves as a recipe to leverage functional correspondences for zero-shot behavior transfer, where functional correspondences are guided by an expert. 

%===============================================================================

\section{Related Works}

\para{Test-time Policy Interventions} Runtime policy interventions are a policy failure mitigation, where-in the policy's execution is intervened and new knowledge is supplied in-order to help mitigate the failure. For instance, a line of work directly proposes interventions on the policy's behavior space, steering the policy into desired modes either through human feedback~\cite{wang2024inference}, through Q functions optimized on large scale offline datasets~\cite{nakamoto2024steering}, or through predictive modeling~\cite{wu2025foresight, qi2025strengthening}. Another line of work proposes intervention directly on the policy's observations, with synthesized observations to remedy known causes of failure~\cite{hancock2024run}. Our work also proposes policy observations with functionally similar ID observations to generalize to novel out-of-distribution conditions. 

% \abnote{Interventions directly on the policy's observation space: BYOVLA. You can also intervene on the output of the polic, i.e. in the behavior space. This is sometimes called policy steering and can be done via human feedback \cite{wang2024inference} 

\para{Functional Correspondence for Behavior Transfer}
The ability to transfer behaviors from one set of objects to an unseen set of objects hold the potential to unlock robot generalization in the wild. This problem has been studied through functional correspondences~\cite{lai2021functional}. Prior work have leveraged functional correspondences to directly transfer behaviors across objects from a single demonstration to novel objects in a one shot manner, or in a zero shot manner by leveraging an affordance dataset ~\cite{tang2025functo, liu2024one, ju2024robo}. Here, functional correspondences are typically established through keypoint based reasoning. In our work, we establish functional correspondences through a correspondence description provided by an expert. Furthermore, instead of directly adapting retrieved behavior, we intervene using the functionally similar training observation

% \para{Approaches to OOD Generalization in Imitation Learning:} 
% Generalization to OOD conditions has been a central challenge in imitation learning. OOD conditions can stem from a multitude of reasons, including covariate shift, new object configurations and new environment conditions. A foundational approach to mitigating this challenge has been DAgger\cite{ross2010efficient}, which requests corrective demonstrations from the expert to augment the training data and retrain. Collecting more demos and retraining can be expensive. In our work, we seek feedback from the expert to establish function correspondences between ID and OOD scenes to repurpose ID behaviour under OOD conditions.

%===============================================================================

\section{Problem Formulation}
\label{sec:psa}

\para{Environment, Observation, \& Action Models}
% Policy, observation
We model the robot's environment $\env \in \envSpace$ as broadly consisting of factors external to the robot such as the objects in the scene, the background, camera configurations,  etc. 
In a particular environment $\env$, the robot senses its proprioceptive states $q \in \mathcal{Q}$ (e.g., end-effector pose, gripper state) and uses a sensor $\sigma : \stateSpace \times \env \to \mathcal{I}$ to obtain high-dimensional RGB image observations of the scene. 
At any time $t$, let the stacked image-proprioception observations be, $\obs_t \in \obsSpace := \mathcal{I} \times \mathcal{Q}$. 
Finally, let $\action \in \actionSpace$ be the robot's action (e.g., end-effector positions and rotations and gripper action). 

% The policy takes in as input these observations $ \obs \in \obsSpace := \mathcal{I} \times \mathcal{Q} $, which includes RGB image data $ I \in \mathcal{I}$ and the proprioceptive states $q \in \mathcal{Q}$ (e.g., end-effector pose, gripper state), 
\para{Training Data} For training the visuomotor policy, we assume access to a dataset of observation-action tuples, $\IDDataset := \{(\obs^i_t, \action^i_t)\}^N_{i=1}$,  drawn from a set of $M$ environments we treat as ``in distribution'': $\IDEnv := \{\IDEnv^1, \IDEnv^2, \hdots, \IDEnv^M\}$. For example, a training distribution of environments could consist of $M$ unique objects and their configurations in the environment. 
% \abnote{Give examples of what the environments could be and how they could influence the observation-action data.} 

\para{Visuomotor Policy} 
Let the robot's policy be a multimodal imitative action generation model \citep{chi2024diffusionpolicy,leebehavior} denoted by $\pi(\actiontraj_{t} \mid  \obstraj_{t})$. 
Here $\actiontraj_{t} := \action_{t:t+T}$ is a $T$-step action plan and $\obstraj_{t} := \obs_{t:t-H}$ is an $H$-step history of observations. 
We assume that the policy network first encodes any observation into a corresponding latent state, $\latent_t = \enc(o_{t})$, via an encoder. 
% predicts the robot's t step action plan denoted by $a_{t} := a_{t:t+T}$ with each action in the sequence specifying end-effector positions and rotations. 
% Note $\obs_{t} := \obs_{t:t+T}$ denotes the observation sequence, 
Let $\latenttraj_{t} = \enc(\obs_{t:t-H})$ be a sequence of latent state embeddings.
The policy is pre-trained via an imitation learning loss on the in-distribution dataset of observation-action pairs from $\IDDataset$.

% The robot's visuomotor policy is trained using behavior cloning (BC) with a dataset of expert trajectories. 
% Let $\IDEnv$ be the robot's training environment. Expert trajectory $\ExpertTrajectory$ collected in $\IDEnv$ is a sequence of observation ($\obs_{t}$), action ($a_{ t}$) pairs, where observations are drawn from the environment, $\obs_{t} \in \stateSpace \times \IDEnv$. 
% The training or in-distribution (ID) dataset a collection of all such action, observation pairs, denoted by $\IDDataset := \{ (\obs_{i, t}, a_{i, t}) \}_{i, t = 0, 0}^{i, t = N, T_{i}}$. 
% Here $N$ refers to the total number of expert trajectories in the dataset, and $T_{i}$ denotes the number of timesteps in each trajectory ($\ExpertTrajectory_{i}$) 
% We denote the set of all ID observations as $\IDObsSet$. 
% With BC, $\pi$ is trained under a MLE objective as defined in Eq.~\ref{eq:bc}. This enables $\pi$ to mimic actions corresponding to in-distribution observation $\obs_{t} \in \IDObsSet$. 

% \begin{equation}
%     \mathcal{L}_{BC} = \E_{(\obs_{i, t}, a_{i, t}) \sim \IDDataset} \| a_{i, t} - \pi (\enc (\obs_{i, t})) \|^{2}
%     \label{eq:bc}
% \end{equation}

\para{OOD-to-ID Generalization via Functional Correspondances} 
Given a visuomotor policy $\pi(\actiontraj_{t} \mid \obstraj_{t})$ pre-trained on behaviors from in-distribution environments $\IDEnv$, we seek to generalize the robot's task performance to \textit{out-of-distribution} (OOD) environments,  $\OODEnv$. 
Since the general problem of OOD generalization is an extremely challenging open problem, in this paper we assume that (1) $\IDEnv$ and $\OODEnv$ differ only by the objects present in the scene and background color (but the environment geometry remains the same), 
% \abnote{is it background color? the background scene geometry?}
(2) the training observations $\IDObsSet$ and deployment time observations $\OODObsSet$ are obtained on the same robot embodiment, and (3) we have access to the training data, $\IDDataset$. 

% Given a BC visuo-motor policy $\pi$, pre-trained on observations $\IDObsSet$ drawn from the $\env_{ID}$, our goal is to improve it's performance under observations $\OODObsSet$ drawn from $\OODEnv$.
Our key idea is to identify \textit{functional correspondences} between the test-time OOD scene---in which the base policy would fail to act correctly---and training-time ID scenes, in which the policy can generate high-quality behaviors. Functional correspondences identify parts of the image observations with similar affordances for the task at hand. Intuitively, learned robot behaviors should be transferable across observations whose affordance maps are aligned, i.e., observations where regions that have similar affordances overlap. Thus, we aim to retrieve ID observations whose functional correspondences are aligned with the test-time OOD observation.

\para{Problem Formulation: Expert-Guided Functional Correspondences} 
The core challenge lies in identifying the functional correspondences across the OOD image observations and the ID image observations. Humans possess the ability to infer object affordances, and generalize them to novel objects. Thus, we propose to leverage experts feedback in the form of natural language to acquire these functional correspondences between the OOD and the ID image observations.

Formally, let the \textbf{functional correspondance map} be denoted by $\Phi : \mathcal{I} \times \mathcal{I} \times \mathcal{L} \to \mathcal{P}(\Omega \times \Omega)$.  Given two images $i, \hat{i} \in \mathcal{I}$ and a natural language description $l \in \mathcal{L}$ provided by the expert, this mapping returns all pairs of functionally corresponding image segments $(\omega, \hat{\omega})$ where $\omega \in \Omega$ are image segments from image $i$ and $\hat{\omega} \in \hat{\Omega}$ are image segments from image $\hat{i}$. 
Here, $\mathcal{P}(\Omega \times \Omega)$ is the powerset of all paired image segments. 
Let $K$ be the number of corresponding image segments. Thus, 
% , $\omega \in \Omega$ and $\hat{\omega} \in \hat{\Omega}$, one from each image observation, which functionally correspond based on the features described in $l$. 
the functional correspondence map is defined as: 
% Here, $\mathcal{L}$ is the space of language inputs, $\Omega$ is the space of all segments of the image observations, and $\mathcal{P}(\Omega \times \Omega)$ is the powerset of all pairs of image segments.
\begin{equation}
    \Phi(i, \hat{i}, l) := \{ (\omega_j, \hat{\omega}_j) \mid j \in \{0, 1, \ldots,  K\} \}  
    \label{eq:functional-map}
\end{equation}
% Here, $i_{t}$ and $\hat{i}_{t}$ are the ID and OOD image observations respectively, $l_{t} \in \mathcal{l}$ is the expert language input describing the functional correspondences across the OOD and ID images and ($\omega^{i}, \hat{\omega}^{i}$) are a pair of functionally corresponding the subregions from $i_{t}$ and $\hat{i}_{t}$, described by $l_{t}$

We measure the \textbf{functional correspondence alignment} of any two images via $\funccor : \mathcal{P}(\Omega \times \Omega) \rightarrow \mathbb{R}$. 
In this work, we model $\funccor$ as the total Intersection over Union (IoU) between functionally corresponding regions of the images returned by the functional correspondance map, $\Phi$:
% We measure the \textbf{functional correspondence alignment} of any two images via $\funccor : \mathcal{P}(\mathcal{M} \times \mathcal{M}) \rightarrow \mathbb{R}$. Specifically, we model $\funccor$ as the total Intersection over Union (IoU) between functionally corresponding regions of the images returned by the functional correspondance map, $\Phi$:
\begin{equation}
     \funccor(\Phi(i, \hat{i}, l)) = \sum_{i=0}^{K} \text{IoU}(\omega_{j}, \hat{\omega}_{j})    
     \label{eq:functional-alignment}
\end{equation}
% \pranay{$\Omega$ as both space and the segmentation mask is confusing, also not $\hat{\Omega}$ is undefined. Change symbols for space of segmentation masks. Have used $\mathcal{M}$ in method for space of image segmentations. Think about how to change} 
% \abnote{Yeah, agreed. I think we should use a different notation for all sets of image segments vs. the set of segments for an image.}
Finally, given any OOD observation $\obsood = (\hat{q}, \hat{i})$ observed by the robot at deployment time and an expert language input $l$ describing the functional correspondences, we retrieve the ordered set of in-distribution observations $\FuncCorrIDObsSet = (o_1, o_2, \dots, o_k) \subseteq \IDObsSet$ ranked by their functional alignment from Eq.~\eqref{eq:functional-alignment}. 
% $ \funccor(i_1, \hat{i}, l, \Phi) \leq  \dots \leq \funccor(i_k, \hat{i}, l, \Phi)    
For intervention, we use the behaviors extracted from the top-$M$ observations in $\FuncCorrIDObsSet$.

\begin{figure}[t!]
    \centering
    \includegraphics[width=\textwidth]{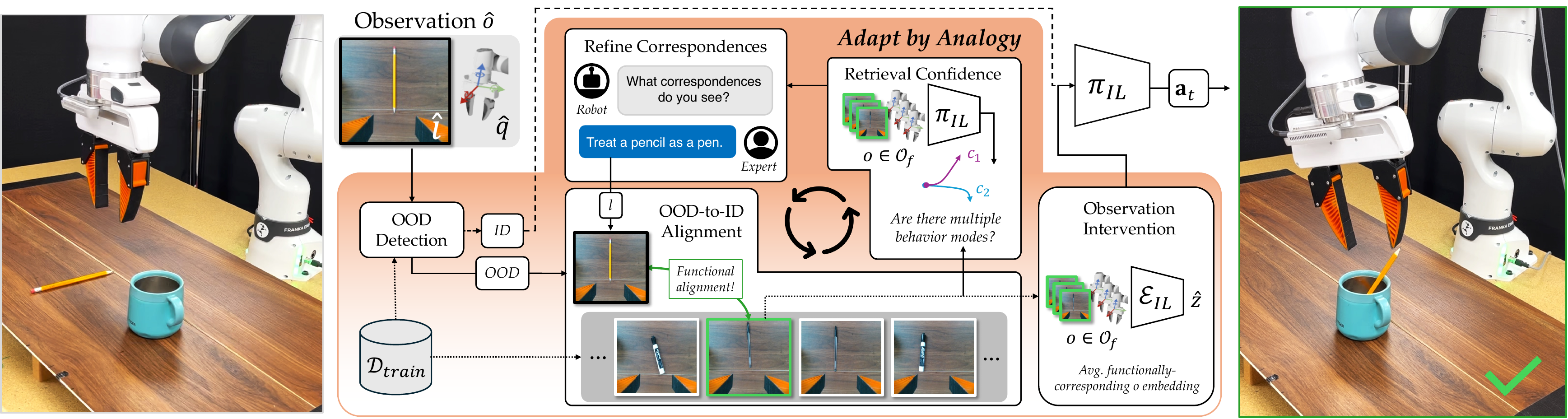}
    \caption{\textbf{Adapting by Analogy} consists of four key phases. (left) First, we run a fast OOD detector by checking the cosine similarity between the current observation $\hat{\obs}$ and the training observations. (center, top-left) Given a correspondence description $l$, we establish OOD-to-ID functional correspondences to retrieve corresponding ID observations (center, bottom). We refine the correspondances with the expert as long as there is ambiguity in the predicted behavior mode (center, top-right). Once finalized, we intervene on the observations and execute the planned actions (right). }
    \vspace{-1em}
    \label{fig:method_figure}
\end{figure}

\section{Method: Adapting by Analogy}
\label{sec:method}

\subsection{Detecting Out-of-Distribution Observations}
\label{subsec:ood_det}
% The first step is to detect whether the current test observation $\hat{\obs}$ is OOD. 
% We assume that $\pi$ only performs reliably with $o_{t} \in \mathcal{O}_{ID}$, and use OOD detection as a proxy to check whether we can rely on policy predictions.
At each timestep, our method first detects if the robot's observations $\obsood$ are anomalous via a fast OOD detector. We measure the cosine similarity between the encoded observation $\hat{z} = \enc(\obsood)$ and embeddings of in-distribution observations $z \in \enc(\obs_i), ~\obs \in \IDObsSet$ via $\text{IDScore}(\hat{\obs}, \IDObsSet) := \min_{\obs \in \IDObsSet}  \frac{\enc(\obsood) \cdot \enc(\obs)}{\| \enc(\obsood) \|\| \enc(\obs) \|}$.
% Note that encoder $\enc : \mathcal{O} \to \mathcal{Z}$, maps the robot observations ($\obs$) to the embedding space $\mathcal{Z}$
% \begin{equation}
%         \text{IDScore}(\hat{\obs}, \IDObsSet) := \min_{\obs \in \IDObsSet}  \frac{\enc(\obsood) \cdot \enc(\obs)}{\| \enc(\obsood) \|\| \enc(\obs) \|} 
%     \label{eq:eq_OOD}
% \end{equation}
If the IDScore 
% obtained via Eq.~\ref{eq:eq_OOD} 
is above a threshold $\lambda$, then we deem the observation to be nominal and directly execute the action $\hat{a} \sim \pi(\cdot \mid \obsood)$. 
Otherwise, we deem the observation to be OOD, 
% ask the user for an initial description of , and intervene on the observation before action generation via functional correspondences.
ask the expert for an initial language instruction $l$ describing relevant functional correspondences, and use those to intervene on the observation before action generation. 

% functionally aligned ID observations $\FuncCorrIDObsSet$.

% \subsection{Learning and Matching OOD-to-ID Functional Correspondences}
% Given, the OOD observation $\hat{o} = (\hat{q}, \hat{i})$, the ID observations $\obs = (q, i) \in \IDObsSet$, and the expert language input $l$, we first extract the set of functionally corresponding image segments using the \textbf{functional correspondence map} $\Phi$ (Eq.~\ref{eq:functional-map}). 
 % $\Phi$ computes semantic segmentation masks $\Omega, \hat{\Omega}$ of the image observations $i, \hat{i}$ respectively, and extracts functional correspondence features from $l$. 
\subsection{Establishing OOD-to-ID Functional Correspondences}
\label{subsec:func_corr_matching}

Given an OOD observation $\hat{o} = (\hat{q}, \hat{i})$ identified via our fast anomaly detector and the expert's language description $l$, we want to intervene on the policy by reusing learned behaviors from functionally similar ID observations. 
This requires computing the functional alignment from Eq.~\ref{eq:functional-alignment} between $\hat{o}$ and every ID observation $\obs \in \IDObsSet$. 
However, implementing this matching is challenging in practice for two reasons: first, it is computationally expensive (requiring image segmentation and IoU computation over all corresponding segments), and second, matching correspondances between 2D image segments does not directly reveal correspondences in the high-dimensional robot state. 
Thus, we filter the demonstration dataset $\IDDataset$ consisting of $N$ observation-action trajectories $\tau$ to retrieve one observation $\obs \in \tau$ per trajectory which contain similar proprioceptive states $q$ to the test-time robot state $\hat{q}$. 
% first filtering $\IDObsSet$ to retrieve observations where the proprioceptive state of the robot $q$ is within a threshold $\lambda_{q}$ of the test-time robot state $\hat{q}$. Additionally, we only pick the closest observation from each demonstration $\tau := \{(\obs_0, \action_0), \hdots, (\obs_T, \action_T)\} \in \IDDataset$. 
% In addition to reducing the search space, it serves another critical purpose. Images are 2D, and robot behaviors are high dimensional, thus high IoU across the image segments does not always mean functional alignment. By searching around the robot's current state, we ensure that IoU across the image segments means functional alignment. 
Mathematically, for a distance threshold $\lambda_q \in \mathbb{R}^+$ and the current configuration $\hat{q}$, let the filtered observation dataset $\mathcal{O}_{q} \subset \IDObsSet$ be:
\begin{equation}
    \mathcal{O}_q = \left\{ \obs \,\middle|\, \obs = \arg\min_{(o, a) \in \tau} \big( \| q - \hat{q} \|_2 - \lambda_{q}\big), \quad \forall \tau \in \IDDataset  \right\}
\end{equation}
% Here $N$ is the total number of expert trajectories in the dataset $\mathcal{D}_{ID}$. 

Using this filtered dataset, we can now compute our \textbf{functional correspondence map} from Eq.~\ref{eq:functional-map}
% $\hat{i}$ and $i \in \mathcal{O}_{q}$.
% We use $\funccor$ to retrieve ID observations $\FuncCorrIDObsSet$. 
% Our functional correspondence map is instantiated 
via two internal models: one which converts the expert's language feedback $l \in \mathcal{L}$ into a functional feature set (denoted by $\phi_l$) and another which semantically segments each ID image $i \in \mathcal{O}_{q}$ and the current OOD image observations $\hat{i}$ to generate the set of image masks and semantic labels denoted by $\hat{\Omega}$ and $\Omega$ respectively. We use Grounded Segment Anything~\cite{ren2024grounded} for semantic segmentation. 
% , we run Grounded Segment Anything \cite{SAM} on the OOD and ID image observations, $\hat{i} \in \hat{\obs}$ and $i \in \obs$, to obtain  

% that allow either coarse- or fine-grained matching.  
% We next use the language input to generate features for correspondence matching.
Next, the expert's language input $l$ is decoded into a set of correspondence features $\phi_l$ that can be applied to the semantic segmentations $\Omega, \hat{\Omega}$ to return a set of $K$ functionally corresponding image segments $(\omega^j, \hat{\omega}^j), j \in \{0,\hdots,K\}$. 
% For example, we can match image segments with different semantic labels (e.g., \textit{``Match pencils with pens''}) or have specific regions of the segments which correspond (e.g., \textit{``Align the right/left edge of the segmentation masks''}). 
For example, the $\phi_l$ that is decoded from $l=$\textit{``Match pencils with pens''} lifts pixels corresponding to the segmentation label \textit{`pencil'} in the OOD image, and pairs it with pixels corresponding to the label \textit{`pen'} in the ID images. 
In this work, we use a templated $l$, but future work could explore the use of LLMs as an interface. 
% Overall, the functional correspondence map is implemented as
% \begin{align}
%     \Phi(i, \hat{i}, l) := \phi_l(\Omega, \hat{\Omega})  
% \end{align}

Ultimately, after $\Phi$ extracts the set of functionally corresponding image segments, we measure their alignment using Eq.~\ref{eq:functional-alignment}. Each ID observation $o \in \mathcal{O}_{q}$ is ranked based on its functional alignment with the OOD observation $\hat{o}$ to obtain the ordered set of functionally corresponding ID observations $\mathcal{O}_{f} \subseteq \mathcal{O}_{q}$ used during intervention (Sec.~\ref{subsec:intervention}). 

\subsection{Refining Functional Correspondences Until Confident} 
\label{subsec:action_mm}
Thus far, we have assumed that the initial expert description $l$ of functional correspondences was sufficient for the entire task. 
However,  
% The retrieval of functionally aligned observations $\mathcal{O}_{f}$ relies on an expert feedback $l$. 
 correspondences may evolve during task execution. 
For example, consider the task of picking up trash and sorting it into organic and recycling. 
The functional correspondence between types of trash (organic and recycling) does not matter initially when the robot is planning a grasp, but becomes relevant once the item has been picked up and needs to be sorted. 
Thus, our method interactively refines the functional correspondence description until the robot is confident in the behavior it has retrieved. 
% Thus, it is critical to detect the need for new functional correspondences during execution and solicit $l$ accordingly.
% Based on our original insight that learned behaviors are transferrable across functionally aligned observations, we infer that functionally aligned observations ($\obs_{t} \in \IDObsSet$) must exhibit similar behaviors ($a = \pi(\obs)$). 
    % \item Functionally \textit{misaligned} observations result in diverse robot behaviors. 
    % % \item Thus as we filter for functionally corresponding observation the behavior modes the policy can exhibit should decrease. We measure this difference to as a proxy for the robot's confidence in the current set of retrieved functionally corresponding samples. 

Intuitively, a well-established functional correspondence will reduce the diversity in robot action plans, focusing on the ``correct'' behavior mode. 
% So we measure how many behavior modes get filtered through correspondence matching. 
% \item for each functionally-similar observation $\obs \in \mathcal{O}_{f}$ given the current $l$, we want an estimate of the visuomotor policy's capabilities in order to understand if there is ambiguity in how the robot should act. If the robot has established a complete functional correspondence, then there should be only one behavior "mode"; if there are multiple distinct behaviors induced by the retrieved observations, then the robot needs to refine the description $l$. 
To quantify the relevant behavior modes before functional alignment, we obtain a set of action plans $\mathcal{A}_{q} := \{ \actiontraj \sim \pi(\cdot \mid \obs) \mid \obs \in \mathcal{Q}_{q} \}$ for all observations with the same proprioceptive state via a forward pass through the policy. Behavior mode labels are obtained by fitting $n_c$ clusters to $\mathcal{A}_{q}$ via K-means clustering. 
Since the current functionally-aligned observations are a subset $\mathcal{Q}_{f} \subseteq \mathcal{Q}_{q}$, we can  obtain labels for all functionally-aligned \textit{action plans} $\mathcal{A}_{f} \subseteq \mathcal{A}_{q}$ and measure the reduction in behavior modes via the entropy over the action plan labels. 
As long as the entropy in the retrieved actions is high, the robot keeps asking the expert to refine their functional correspondence description $l$ by showing them the current observations and their behaviors, then re-doing the OOD-to-ID matching from Sec.~\ref{subsec:func_corr_matching}. 
% plans then the robot requests that the expert augments their functional correspondence description ...
% We request expert feedback and update $\mathcal{O}_{f}$ until the $H(\mathcal{A}_{f})$ is below the threshold $\lambda_{mm}$. 

% \begin{equation}
%     H(\mathcal{A}_f) = -\sum_{i=1}^{n_c} \left( \frac{1}{|\mathcal{A}_f|} \sum_{a_{t} \in \mathcal{A}_f} \mathbb{I}(c(a_{t}) = i) \right) \log \left( \frac{1}{|\mathcal{A}_f|} \sum_{a_{t} \in \mathcal{A}_f} \mathbb{I}(c(a_{t}) = i) \right)
% \end{equation}

\subsection {Intervening on Observations to Generate Functionally-Corresponding Behavior}
\label{subsec:intervention}
Once the correspondence description $l$ is complete and the retrieved action mode uncertainty is sufficiently low, the robot intervenes on its observations to generate functionally ``correct'' behavior. 
% The final step of our method, is to intervene policy action generation using the behaviors from $\mathcal{O}_{f}$. 
Specifically, observations in the final refined $\mathcal{O}_{f}$ are ranked based on their functional alignment as measured by Eq.~\ref{eq:functional-alignment}. 
To smooth out action prediction, we generate the final executed action plan by 
% However the highest ranked observation is rarely an exact match i.e. the functionally corresponding image segments rarely exactly overlap. But, empirically this gap tends to average out if we consider the top-N samples.  
% Thus, instead of generating the action using the single highest ranked observation ($\argmax {f(\obs, \hat{\obs}, \Phi) \mid \obs \in \mathcal{O}_[f]}$), 
interpolating the embeddings of $M$-highest ranked corresponding observations: $\hat{z} := \frac{1}{M} \sum_{o \in \mathcal{O}_f} \enc(o)$ before passing the average embedding to the policy network. 

\section{Hardware Experiments}
\label{sec:hardware-experiments}

We conduct a series of experiments in robot hardware to study:
(1) How much does \textit{Adapting by Analogy} improve the visuomotor policy's closed-loop performance on OOD environments induced by novel objects and backgrounds conditions?, 
(2) What kind of features (e.g., base policy's embedding, DINOv2 \cite{oquabdinov2}, or functional correspondences) maximally help observation interventions?, 
(3) How efficient is our method at seeking expert feedback for adaptation in OOD environments?, 
(4) When intervention schemes succeed, are they retrieving functionally-aligned observations?

\para{Real Robot Setup} We use a Franka Research 3 robotic manipulator equipped with a 3D printed UMI gripper \cite{chi2024universal} for our real-world experiments. The RGB image observations $i \in \mathcal{I}$ come from a wrist mounted RealSense D435 camera and a third-person Zed mini 2i camera overlooking the workspace. The overall robot observation $\obs := (i, q)$ consists of the concatenated images and the robot proprioception. More details about our setup can be found in the Supplementary \secref{suppl:setup}.

\para{ID Environments \& Tasks} We train two visuomotor policies on two different real-world manipulation tasks. 
The first task is \textbf{sweep-trash}, wherein robot must sweep trash towards different goals, based on whether the trash is organic and recycling. 
% For evaluation, we divide the task in two sub-goals (A) properly aligning the wiper with the trash, (B) sweeping to the correct location. 
% The training environments are $\IDEnvLabel{Trash} := \{\IDEnv^{\text{paper}}, \IDEnv^{\text{M\&Ms}}\}$
The next task is \textbf{object-in-cup}, where-in a robot arm is tasked with picking up a object such as a marker or a pen and dropping it in a mug. Pens---which are grasped above their center-of-mass---need to be dropped into the mug from the bottom, and markers---which are grasped below their center-of-mass---need to be dropped from the front. We divide the task in 3 sub-goals (A) grasping the object, (B) picking the correct behavior mode based on the grasp, and (C) dropping object into the cup.
% The training environments are $\IDEnvLabel{object} := \{\IDEnv^{marker}, \IDEnv^{pen}\}$. 

\para{Visuomotor Policy Training} We use a diffusion policy \cite{chi2024diffusionpolicy} as the base visuomotor policy $\pi(\actiontraj \mid \obstraj)$. It takes as input $\obs$ and predicts a T-step action plan, where $T = 16$. For \textbf{sweep-trash}, the training dataset $|\IDDataset| = 100$ consists of $50$ demonstrations cleaning up crumpled paper (recycling trash) and $50$ demonstrations cleaning up M\&Ms (organic trash).
For \textbf{object-in-cup}, the policy is trained on $|\IDDataset| = 200$ demonstrations with $100$ placing a marker (dropped from the back) and $100$ placing a pen (dropped from the top).

\para{OOD Environments}  We test on two in-distribution environments and five OOD environments for each task. In addition to pens and markers, we e evaluate sweep trash with one background variation (workspace covered with black cloth), and three novel instances of trash (doritos, crumpled napkin, thumb tacks).  
For the object-in-cup, we test with in-distribution objects, one novel background (workspace covered with black cloth), and three instances of novel objects varying in shapes and sizes (pencil, battery, jenga block). 

% For \textbf{sweep-trash} our OOD environments are $\OODEnvLabel{Trash} := \{\OODEnv^{\text{doritos}}, \OODEnv^{\text{napkin}}, \OODEnv^{\text{thumb-tack}},\OODEnv^{\text{paper-bg}}, \OODEnv^{\text{M\&M-bg}}\}$. 

% For \textbf{object-in-cup} our OOD environments are $\OODEnvLabel{object} := \{\OODEnv^{\text{pencil}}, \OODEnv^{\text{battery}}, \OODEnv^{\text{block}}, \OODEnv^{\text{marker-bg}}, \OODEnv^{\text{pen-bg}}\}$. 

% OOD generalization in both tasks requires reasoning in terms of features that transcend the visual semantics of the objects and thus highlight the importance of leveraging functional correspondences for OOD generalization. 
% Due to the variation in the shapes and sizes of the novel objects, our evaluation showcases how far we can push the idea of OOD generalization by adapting behavior from functionally aligned ID observations. 

\para{Baselines}
We compare our method, \ours, with three baselines. 
% 3 baselines - \textit{Vanilla Policy}, \textit{Policy Embedding}, \textit{Dino Features}. We describe each approach below. \pranay{rename, use macros for colors}
\vanilla is the base visuomotor policy without any intervention mechanism. 
\policyem intervenes on the observations with a similar mechanism to ours, but it retrieves ID observations using cosine similarity in the base policy's learned embedding space, $\enc(\obs) \in \latentSpace$. It does not use any expert feedback. 
% To highlight that BC policies tend to overfit to the training distribution and are not always able to capture the features required to generalize in OOD condition, we compare our method with this baseline.
\dino also intervenes on the base policy, but it retrieves ID samples using cosine similarity in the DinoV2 \cite{oquabdinov2} feature space of the OOD and ID observations. We use this to test if powerful pre-trained vision foundation models can implicitly capture functional correspondences beyond semantic object categories. We use both the class token features and the patch features. 
% : To highlight that vision foundation models, particularly Dino-ViT fail to capture functional correspondences beyond semantic object categories,
For \ours, we generate correspondence features $\phi$ via decoding $l$ into a pre-templated set of features. The choices of the features are
(1) match \textit{`ood object semantic label'} with \textit{`id object semantic label'}, (2) overlap segments of \textit{`ood object'} with \textit{`id object'} (3) Align left/right edge of segments, (4) align top/base of segments, and (5) ``Pass'', which meant that the expert does not want to refine the set of correspondence features. 
% was okay with the multi-modality in the functionally similar behaviors. 
All intervention methods perform matching in the refined set of ID observations based on the robot's current proprioception $\mathcal{O}_{q}$, as described in Sec.~\ref{subsec:func_corr_matching}. 

\para{Evaluation Procedure} 
All methods are evaluated via the same procedure and in the same conditions. For each ID and OOD environmental condition, we perform 10 rollouts of each method, placing the object of interest uniformly at random within a $15$ cm horizontal range on the table. 
With a total of 14 environmental conditions, we collect a total of 140 rollouts for evaluation.

\begin{figure}[t!]
    \centering
    \includegraphics[width = \textwidth]{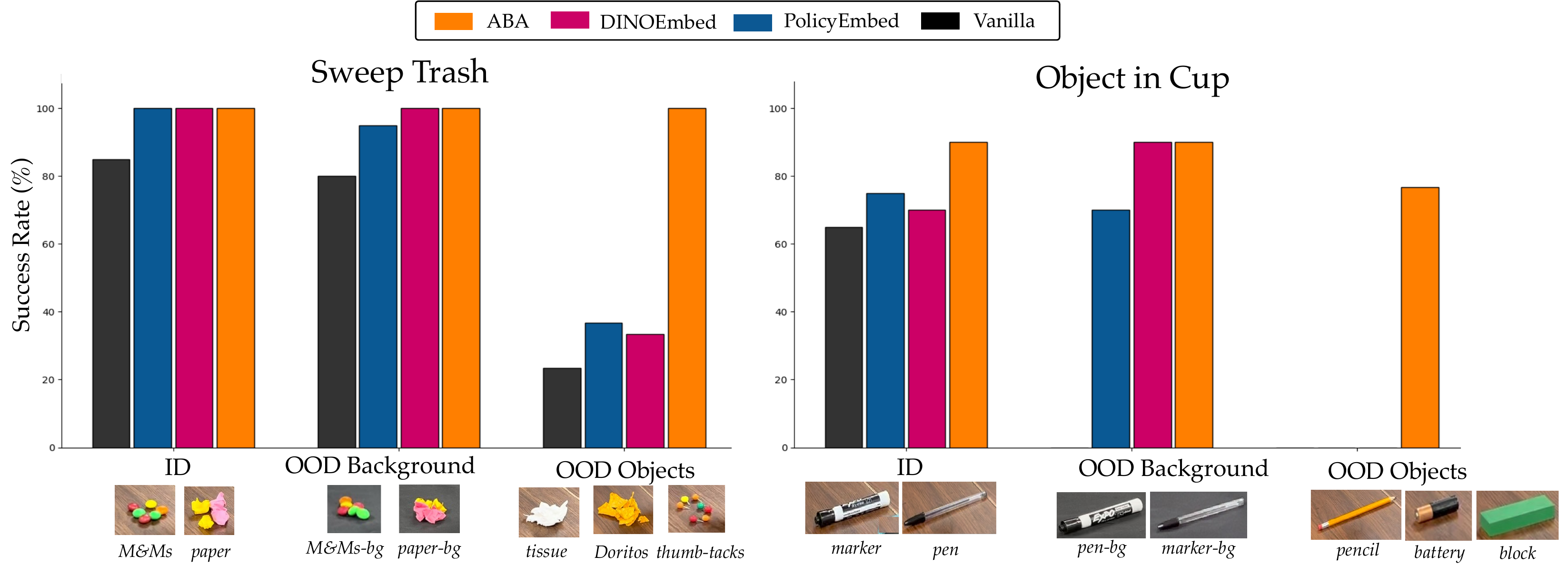}
    \caption{\textbf{Task Success in ID and OOD Environments.} We report the task success rate averaged across 10 rollouts (per each ID and OOD conditions) and averaged across ID, OOD background, or OOD object conditions. For both the sweep-trash and the object-in-cup tasks, we see that \ours consistently achieves the highest task success rate compared to baselines. }
    \vspace{-1em}
    \label{fig:overall_ts}
\end{figure}

\subsection{How much does \ours improve the policy's closed-loop performance?} 
In this section, we compare the overall task success rate of \ours with \vanilla. As shown in \figref{fig:overall_ts}, \ours improves the \vanilla policy even in in-distribution environments by 15\% on the sweep-trash task and 25\% on the object-in-cup task. While the vanilla policy was robust to the novel background for the sweep-trash task, it completely degraded when faced with the novel background in the more challenging object-in-cup task. 
By reasoning about functional correspondences, \ours  improved over the vanilla policy by 20\% on the sweep trash task and by 90\% on the object in cup task, staying robust to task-irrelevant changes to the background. 
Finally, we observe strong OOD generalization with \ours when evaluated under OOD objects where it improves over the vanilla policy by ~76\% on both tasks, showcasing that learned behaviors can be transferred to OOD objects from different semantic categories by reasoning about functional correspondences.  

\subsection{What kind of features maximally help observation interventions?}
In this section, we compare how the features used for retrieving ID observations affect policy performance. For in-distribution environments, on the sweep trash task both \policyem and \dino perform on par with \ours. However, \ours outperforms by ~15\% on the object in cup task. Similar to \ours, both the \dino and \policyem are also robust to the novel background. Interestingly, \dino's performance \textit{improves} under the novel background. We hypothesize that this is due to the exceptional capabilities of the dino features at dense correspondence matching across objects within the same semantic category \cite{oquabdinov2}. 
When tested on OOD objects, both \dino and \policyem struggle, achieving only ~$36.67\%$ and $33.34\%$  success rate respectively on sweep trash. On the object in cup task both baselines failed to successfully complete the task. Taking a closer look at the performance with specific OOD objects revealed common failures at the grasping stage and at picking the correct behavior mode. More analysis in Supplementary \secref{suppl:res}

\begin{figure}[t!]
    \centering
    \includegraphics[width = \textwidth]{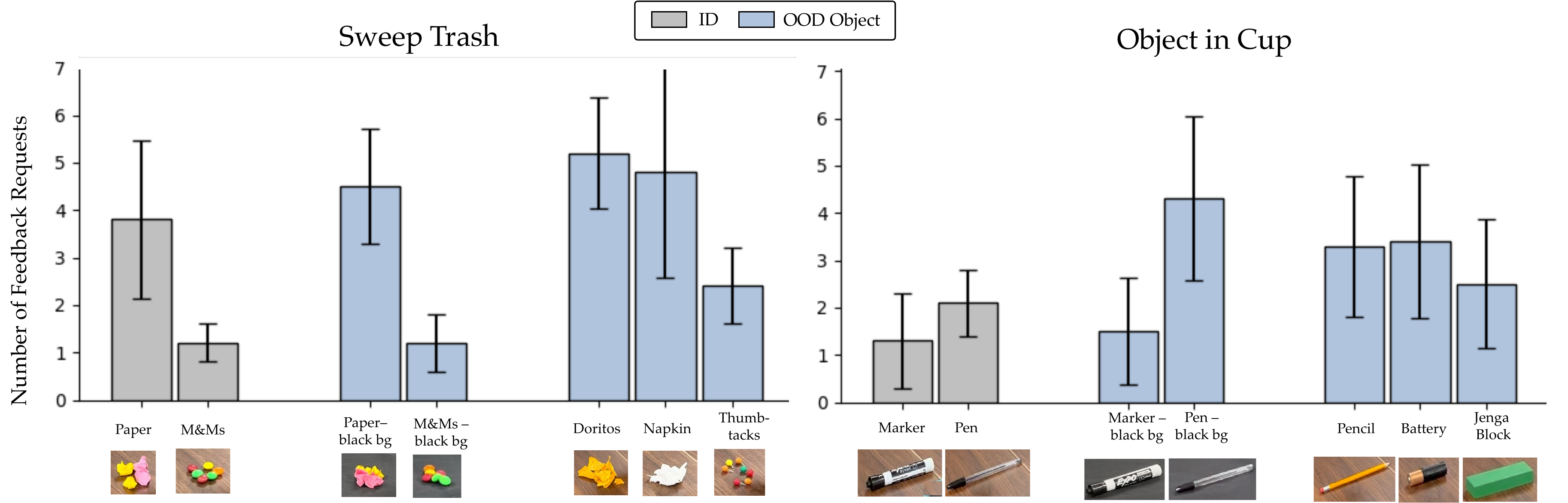}
    \caption{\textbf{Expert Feedback Requested by \ours.} We show mean and standard error for the number of feedback requests across 10 rollouts per each environment. We find that \ours infrequently queries the expert for correspondances, given that sweep-trash has 70 timesteps and object-in-cup has 120.}
    \vspace{-1em}
    \label{fig:feedback_count}
\end{figure}

\subsection{How efficient is \ours at seeking expert feedback in OOD environments?} 

Next we study how often does \ours request feedback from the expert at test-time. \figref{fig:feedback_count} shows the number of times \ours requested feedback on average across 10 rollouts (with standard error bars) for both the sweep trash and the object in cup task, in all the three experiment settings (ID, OOD-Bg, OOD-Object). 

For the \textbf{sweep-trash} task, \ours asks for feedback $3.8 \pm 1.66$ times for ID crumpled paper. In OOD backgrounds,feedback requests increased, e.g., to $4.5 \pm 1.2$ times per rollout for crumpled paper. \ours requested feedback the most for OOD objects, with the highest number of requests for doritos with an average of $5.2 \pm 1.16$ times per rollout. Note that each rollout for sweep trash ran for $80$ timesteps, so this corresponds to asking for feedback ~6\% of the rollout. 
For the \textbf{object-in-cup} task, feedback was requested $2.1 \pm 0.7$ times for the ID pen, and $1.3 \pm 1.0$ times for the ID marker. Similar to sweep-trash, the feedback requests increased with OOD backgrounds: e.g., feedback about the pend was requested $4.3 \pm 1.73$ times per rollout. Finally, amongst the OOD objects, feedback was requested the most for battery at $3.4 \pm 1.62$ times per rollout. Note that each rollout in the object in cup task ran for 120 timesteps.

\subsection{When intervention methods succeed, do they retrieve functionally-aligned observations?} 

\begin{wrapfigure}{r}{0.55\textwidth}
    \centering
    \includegraphics[width=1.0\linewidth]{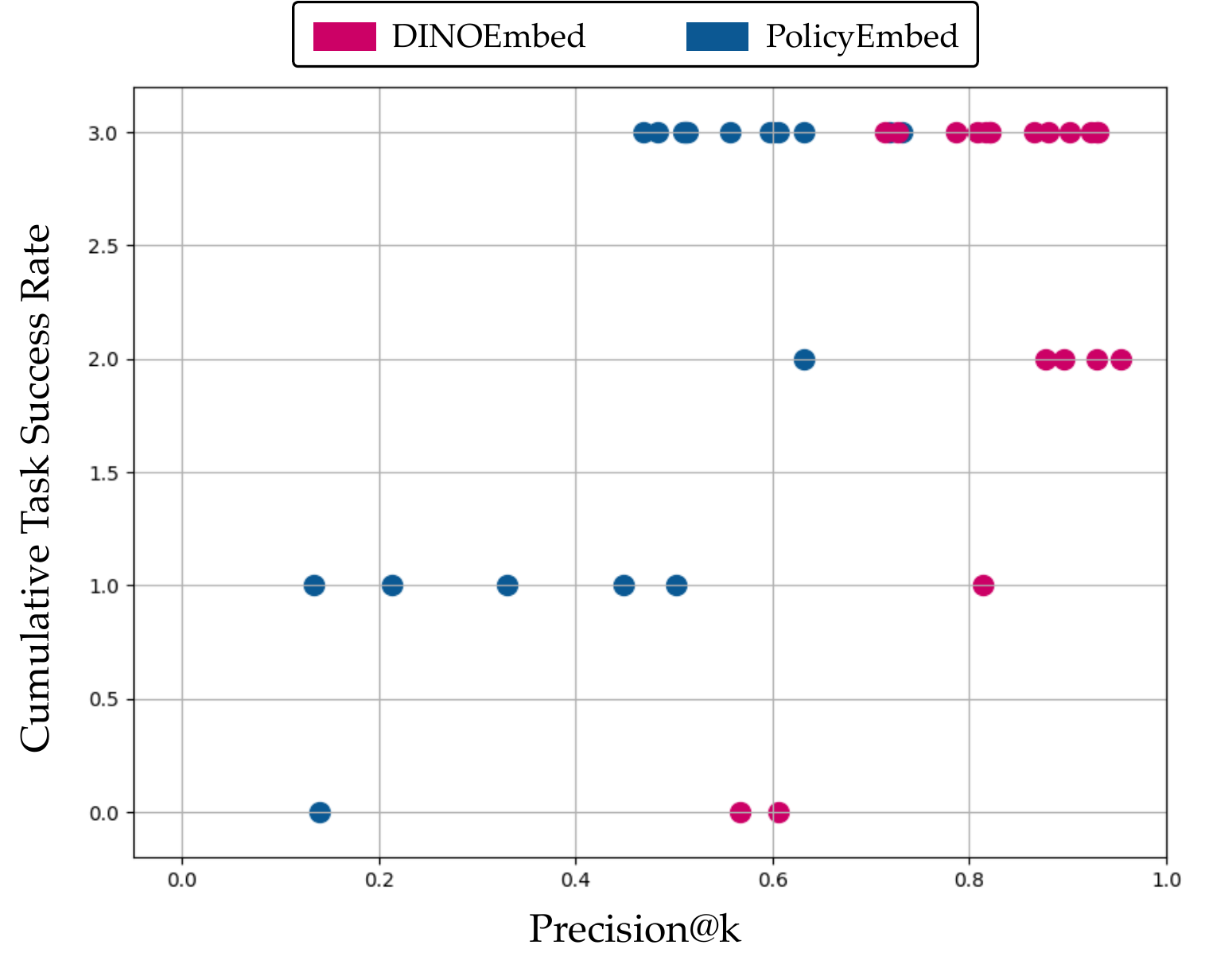}
    \caption{\small{Retrieval overlap with \ours vs. task success. We report precision of the observations retrieved by the intervention based approaches with observations retrieved by \ours, against the task-success for each rollout. The high precision of rollouts with high cumulative task success show that successful rollouts from baselines retrieve observations similar to \ours.}}
    \label{fig:retrieval}    
\end{wrapfigure}

Finally, we checked if retrieving functionally corresponding ID observations was critical for OOD generalization. Thus, we looked at what observations the baselines retrieved when they resulted in successful rollouts. We measured the precision of the set of observations retrieved by the \dino and \policyem baselines against the observations retrieved by \ours. 
% We checked for the presence of functionally corresponding observation in the set of observations retrieved by the baseline during successful rollouts.

Fig.~\ref{fig:retrieval} shows the precision vs. the cumulative success rate, where each dot on the plot is a rollout\footnote{The final precision value is averaged over each timestep where retrieval happened during the rollout.}. The numbers are only for the \textbf{object-in-cup} task in both the ID and OOD background environment configurations, as they had sufficient coverage over successes and failures. Overall, the trends indicate that when the baselines were successful, they also retrieved functionally corresponding observations with high precision. This shows that interventions using functionally corresponding observations help generalize under OOD conditions.

\section{Conclusion}

In this work, we present \textit{Adapting by Analogy}, a method for enabling deployment-time generalization of visuomotor policies by leveraging functional correspondences between out-of-distribution (OOD) and in-distribution (ID) observations. 
Rather than requiring new expert demonstrations for each novel scenario, our approach uses expert-provided functional features---which are interactively refined to represent during task execution---to repurpose existing ID policy behaviors in OOD environments. 
Empirical results across two real-world manipulation tasks with ten OOD environments demonstrate that establishing functional correspondences can improve a diffusion policy's success rate by $~76\%$ to new objects and backgrounds with minimal human intervention.

%===============================================================================

% limtiations do not count towards page limit
\section{Limitations}

Our method is not without its limitations. First, our method assumes that there exists a functional behavior overlap between the OOD scenarios and ID scenarios, enabling the learned behaviors from the ID scenarios to be reused. However, this may not hold in tasks where new objects or environment geometries require fundamentally new strategies that go beyond what was seen in training. Future work should rigorously quantify the robot's \textit{confidence} in retrieving a behavior that is relevant and actively asking for expert demonstrations when no behaviors are relevant. 
Second, our method does still rely on an expert to provide the functional correspondences at test time, which can be challenging for novice end-users and limit the autonomy of the robot. Future work should study the autonomous identification of correspondence features (e.g., via another foundation model). Relatedly, the decoding of the language description into the functional feature set can be ambiguous, prompting the need for future work on grounding natural language into embodied representations. Finally, our system's performance requires a reliable OOD detection mechanism, which should be properly calibrated to balance how frequently the robot has to do interventions and ask for features from the expert. 

%===============================================================================

% \section*{Acknowledgments}

% \clearpage
% % The acknowledgments are automatically included only in the final and preprint versions of the paper.
% \acknowledgments{If a paper is accepted, the final camera-ready version will (and probably should) include acknowledgments. All acknowledgments go at the end of the paper, including thanks to reviewers who gave useful comments, to colleagues who contributed to the ideas, and to funding agencies and corporate sponsors that provided financial support.}

%===============================================================================

% no \bibliographystyle is required, since the corl style is automatically used.
\bibliography{example}  % .bib

%===============================================================================

\newpage
\appendix
\section*{Supplementary}
This is the supplementary material to the paper, Adapting by Analogy: OOD Generalization of Visuomotor Policies via Functional Correspondence. More details can be found on our project page \url{https://anon-corl2025.github.io/project-page/}

\section{Hardware Experiment Setup}
\label{suppl:setup}
\begin{figure}[h!]
    \centering
    \includegraphics[width=\textwidth]{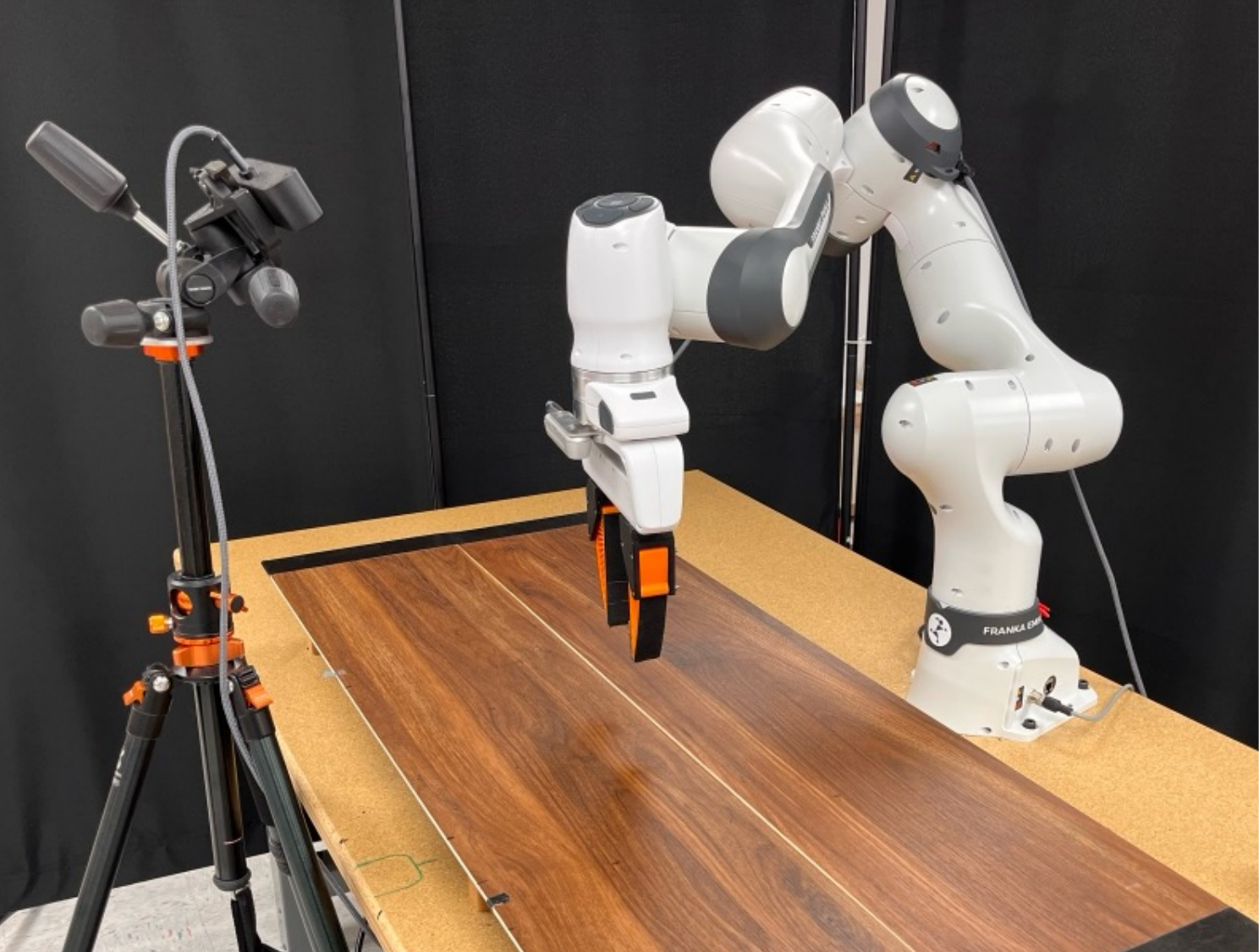}
    \caption{Our hardware experiment setup, we use a Franka Research 3 robot, with a UMI gripper. The RealSense D435 wrist camera, and Zed mini 2i third person camera are placed as shown.}
    \vspace{-1em}
    \label{fig:hardware_setup}
\end{figure}

% \figref{fig:hardware_setup} shows our hardware setup. We employ two cameras, a RealSense D435 camera on the Franka hand and a Zed mini 2i camera placed in front of the
% robot. In order to increase the contact region and compilancy, we replace the original Franka gripper finger with 3D printed gripper finger from \cite{UMI}.

\section{Tasks} 
\label{suppl:tasks}
\begin{figure}[h!]
    \centering
    \includegraphics[width=\textwidth]{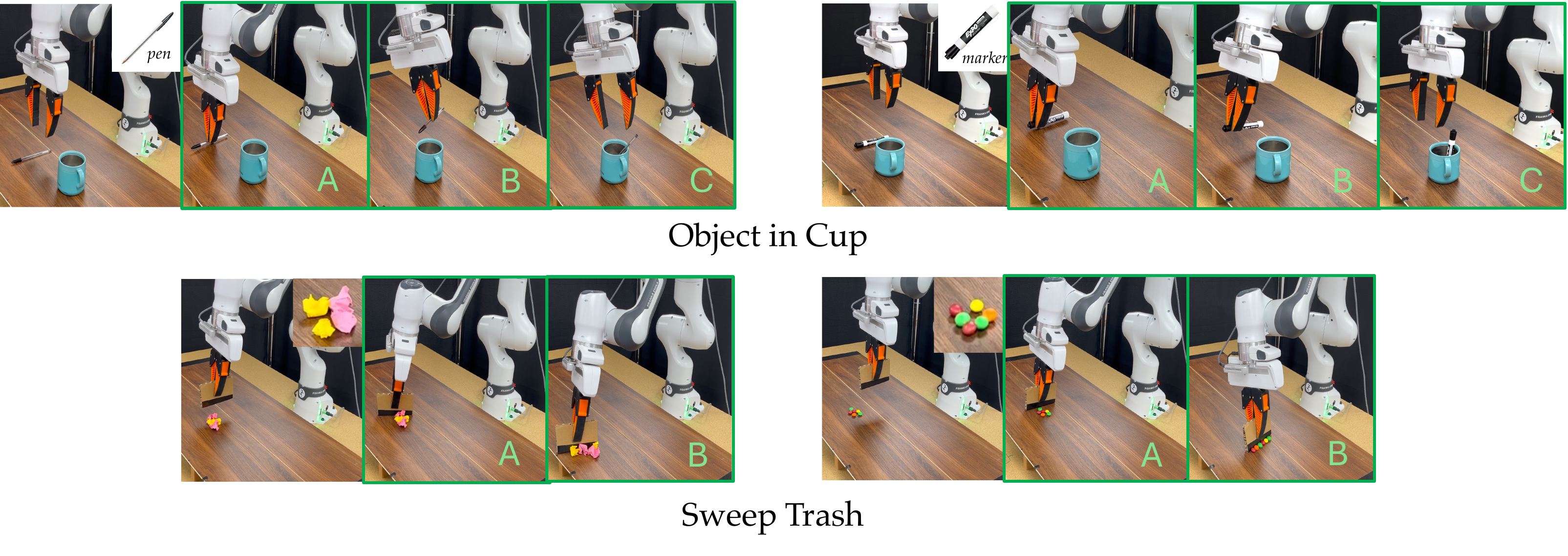}
    \caption{The training demonstrations for our two tasks, with their sub-goals(A, B, C). For the object in cup task, the pen is grasped below the center-of-mass, and is dropped into the mug from the front. The marker is grasped above the center-of-mass and is dropped into the mug from the bottom. For the sweep trash task, paper (i.e., recycling) is swept up, and M\$Ms (i.e., organic) is swept down.}
    \vspace{-1em}
    \label{fig:training_tasks}
\end{figure}

We conduct our experiments on two real-world tasks.
The first task is \textbf{sweep-trash}, wherein robot must sweep trash towards different goals, based on whether the trash is organic and recycling. For evaluation, we divide the task in two sub-goals (A) properly aligning the wiper with the trash, (B) sweeping to the correct location. 
The next task is \textbf{object-in-cup}, where-in a robot arm is tasked with picking up a object such as a marker or a pen and dropping it in a mug. Markers---which are grasped above their center-of-mass---need to be dropped into the mug from the bottom, and pens---which are grasped below their center-of-mass---need to be dropped from the front. We divide the task in 3 sub-goals (A) grasping the object, (B) picking the correct behavior mode based on the grasp, and (C) dropping object into the cup. \figref{fig:training_tasks} demonstrates the various modes and the sub-goals for the two tasks.

\begin{figure}[h!]
    \centering
    \includegraphics[width=\textwidth]{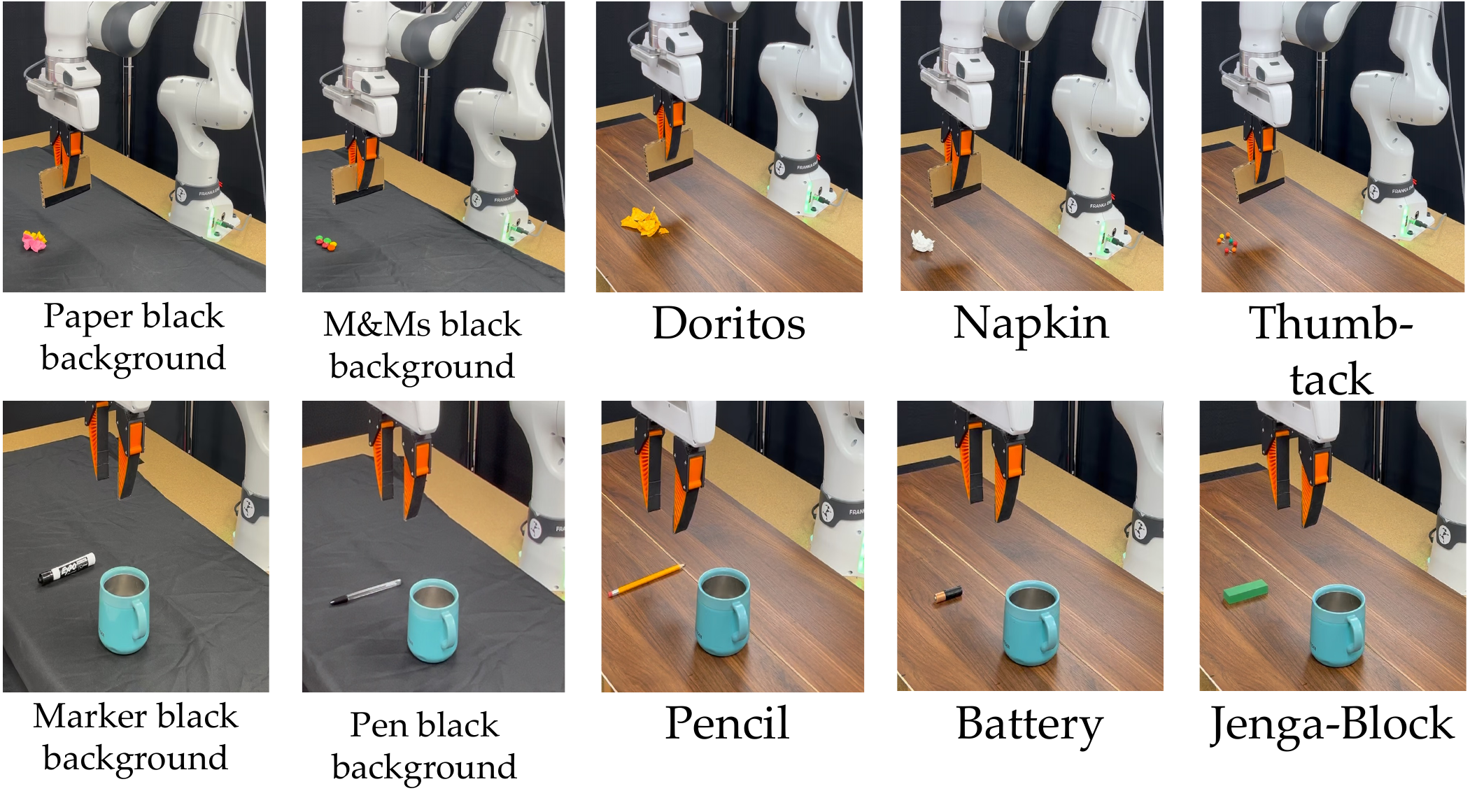}
    \caption{Our OOD environments for both the sweep-trash and object-in-cup task}
    \vspace{-1em}
    \label{fig:ood_envs}
\end{figure}

We evaluate both tasks on the in-distribution conditions and OOD conditions induced by background and novel objects. The OOD environments are shown in \figref{fig:ood_envs} 
For \textbf{sweep-trash} our ID environments are $ \IDEnvLabel{trash} := \{ \IDEnv^{paper}, \IDEnv^{M\&Ms} \} $. The OOD environments are $\OODEnvLabel{Trash} := \{\OODEnv^{\text{doritos}}, \OODEnv^{\text{napkin}}, \OODEnv^{\text{thumb-tack}},\OODEnv^{\text{paper-bg}}, \OODEnv^{\text{M\&M-bg}}\}$. 

For \textbf{object-in-cup} our ID environments are $ \IDEnvLabel{object} := \{ \IDEnv^{marker}, \IDEnv^{pen} \}$. The OOD environments are $\OODEnvLabel{object} := \{\OODEnv^{\text{pencil}}, \OODEnv^{\text{battery}}, \OODEnv^{\text{block}}, \OODEnv^{\text{marker-bg}}, \OODEnv^{\text{pen-bg}}\}$. 

\begin{figure}[h!]
    \centering
    \includegraphics[width=\textwidth]{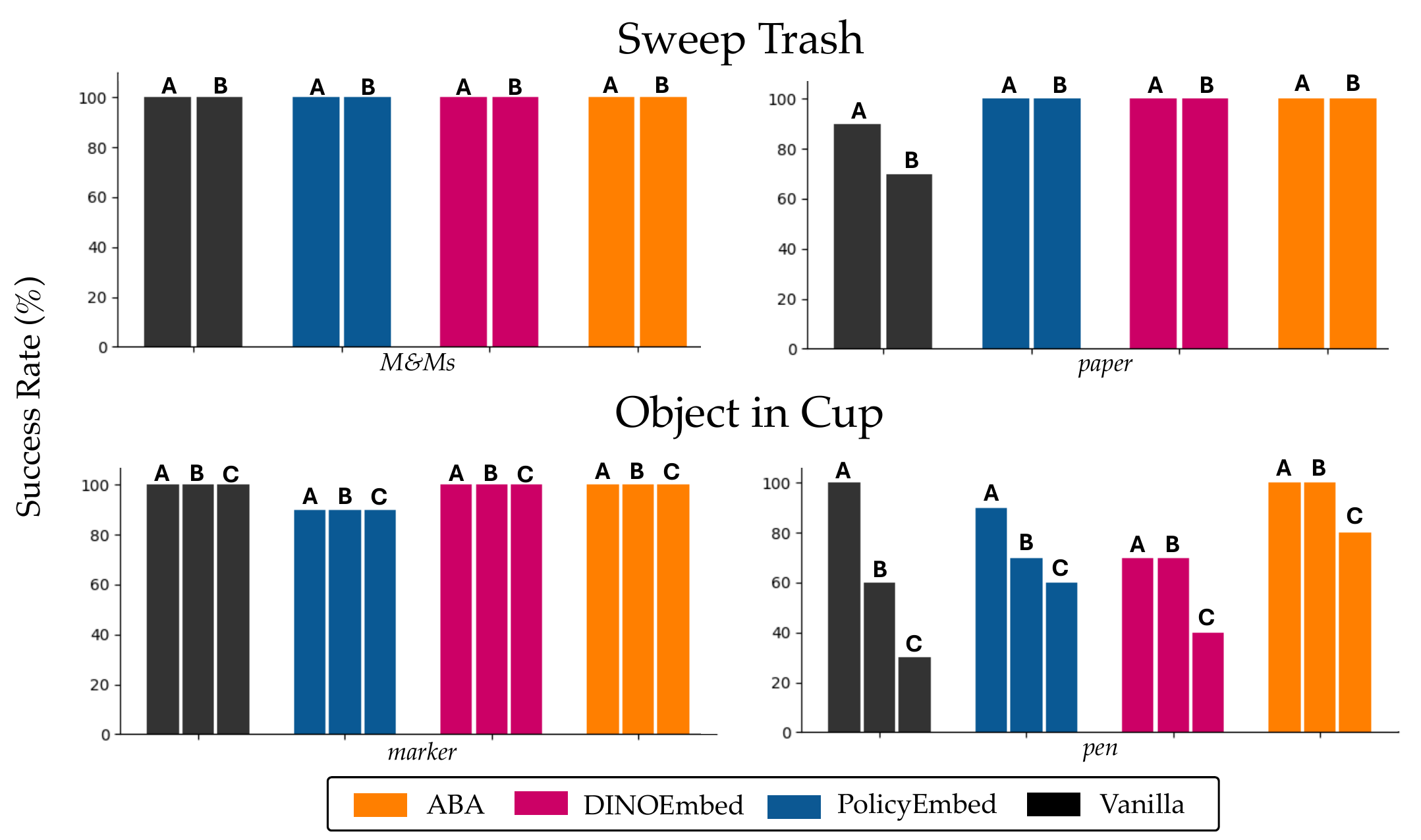}
    \caption{Subgoal Success in each ID Environment. We report the subgoal level task success rate averaged across 10 rollouts. For both the sweep-trash and the object-in-cup tasks, we see that \ours consistently achieves the highest task success rate compared to baselines. }
    \vspace{-1em}
    \label{fig:res_id}
\end{figure}

\begin{figure}[h!]
    \centering
    \includegraphics[width=\textwidth]{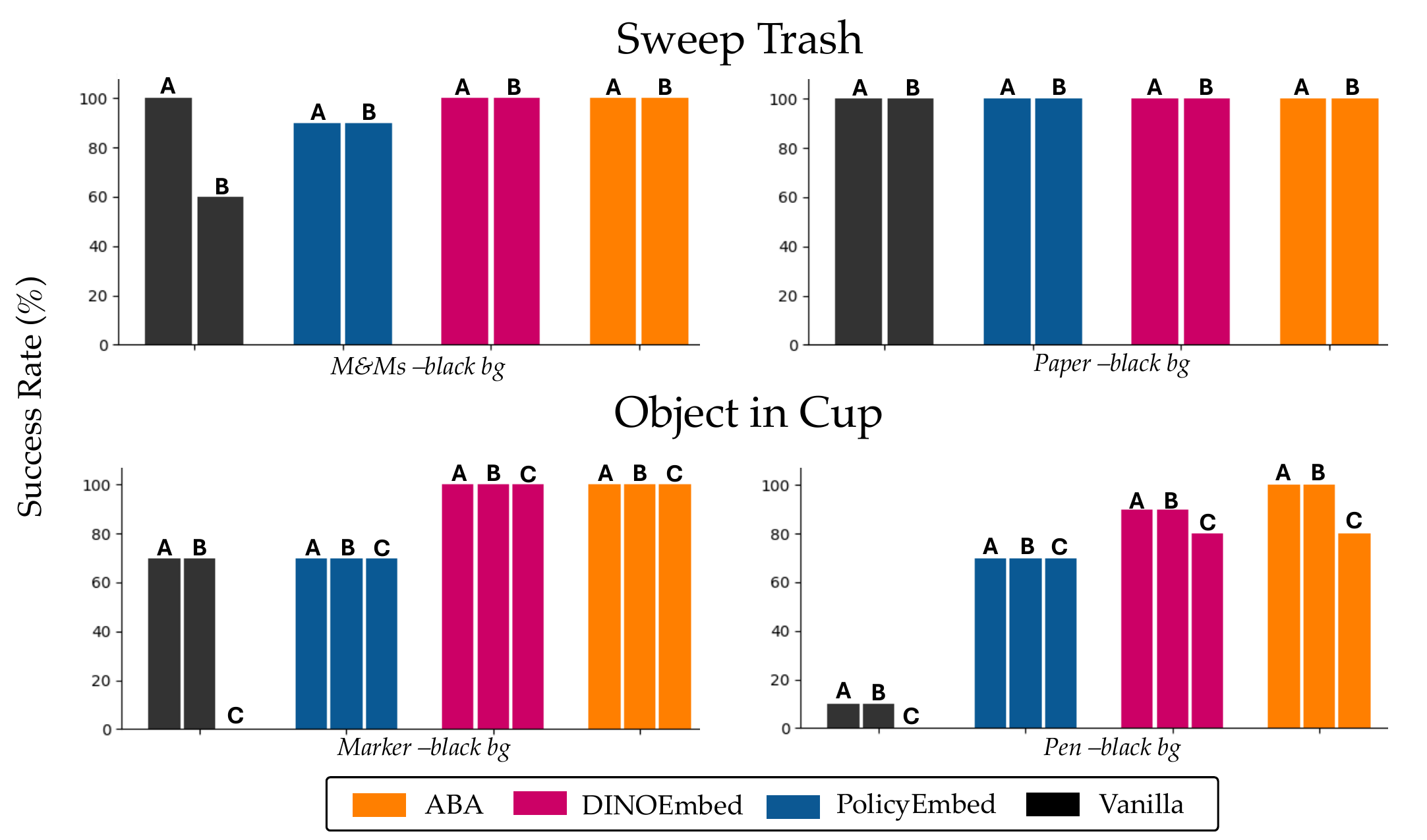}
    \caption{Subgoal Success in each OOD Environment, induced by changing the background. The success rate is averaged across 10 rollouts.  \ours again consistently achieves the highest task success rate compared to baselines.}
    \vspace{-1em}
    \label{fig:res_bg}
\end{figure}

\begin{figure}[h!]
    \centering
    \includegraphics[width=\textwidth]{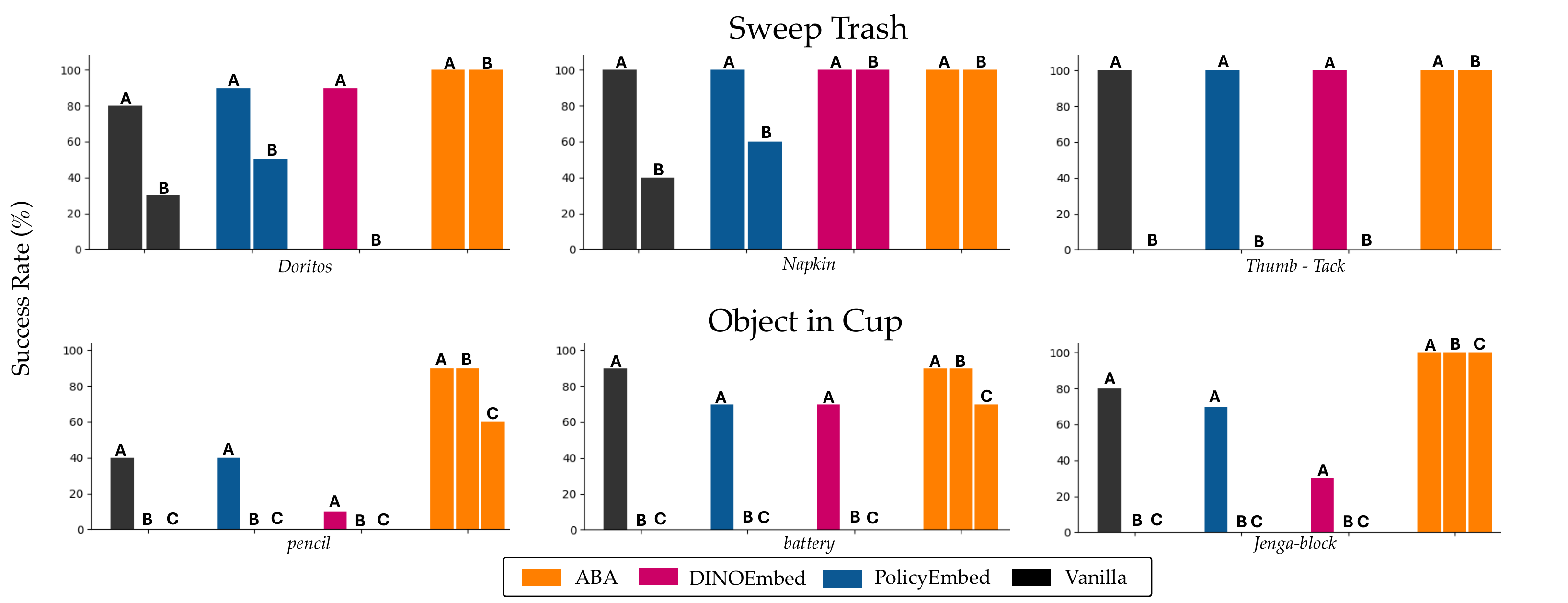}
    \caption{Subgoal Success in each OOD Environment with $3$ novel objects for both sweep-trash and object-in-cup task. The success rate is averaged across 10 rollouts.  \ours again consistently achieves the highest task success rate compared to baselines.}
    \vspace{-1em}
    \label{fig:res_ob}
\end{figure}

\section{Additional Results}
\label{suppl:res}

\subsection{How much does \ours improve the policy's sub-goal level closed-loop performance?}

\figref{fig:res_id} shows that on the sweep-trash task, both \ours and \vanilla are able to successfully accomplish both subgoals on $\IDEnv^{M\&Ms}$. However, in $\IDEnv^{paper}$,  \vanilla fails at aligning the wiper with the paper trash (subgoal A) $10\%$ of the times, and sweeps paper incorrectly (subgoal B) $30\%$ of the times. \ours maintains $100\%$ performance on $\IDEnv^{paper}$.

Showing a similar trend, both \vanilla and \ours show $100\%$ success rate on the $\IDEnv^{marker}$ for the object-in-cup task. On the $\IDEnv^{pen}$, while both \vanilla and \ours are able to grasp the pen (subgoal A) $100\%$ of the times, \ours improves over \vanilla by $40\%$ at picking the right mode (subgoal B), showing a $100\%$ success rate. Finally, since dropping the pen into the cup (subgoal C) is the most fine-grained aspect of the task, both \ours and \vanilla struggle but \ours still improves over \vanilla by $50\%$.

Next, we compare \ours and \vanilla across OOD environments, induced using a novel background ($\OODEnv^{\text{paper-bg}}, \OODEnv^{\text{M\&M-bg}}, \OODEnv^{\text{pen-bg}}, \OODEnv^{\text{marker-bg}}$).

\figref{fig:res_bg} shows that on the sweep-trash task the performance trends are similar to the ID environments, although interestingly instead of $\OODEnv^{paper-bg}$, \vanilla now shows poorer performance on subgoal B of $\OODEnv^{\text{M\&Ms-bg}}$. \ours shows $100\%$ success rate on both goals of both $\OODEnv^{\text{M\&Ms-bg}}$ and $\OODEnv^{\text{paper-bg}}$. On the object-in-cup task, \vanilla struggles on all subgoals of both $\OODEnv^{\text{pen-bg}}, \OODEnv^{\text{marker-bg}}$ environments. \ours improves over \vanilla on both environments showing a $100\%$ performance on all subgoals of both environments, except subgoal C of the $\OODEnv^{\text{pen-bg}}$, where it shows an $80\%$ success rate. This shows that a with novel background, \vanilla fails to even grasp the objects, however interventions with ID observations ignores the OOD conditions induced by the novel background, allowing \ours to uphold closed loop performance under the OOD environments. 

Finally, on OOD environments induced by novel object categories for the sweep-trash task we observe from \figref{fig:res_ob} that while \vanilla is able to align the wiper with the trash, it fails to pick the correct direction for sweeping the trash (subgoal B), as visual features are not enough to decide whether the trash is organic or recycling. \ours is able to successfully accomplish both subgoals for all novel objects as the relevant features for deciding the trash type are supplied by the expert as functional correspondences.

Since the object-in-cup task is more challenging, \vanilla is only performant at grasping (subgoal A). It is able to grasp the pencil with $40\%$, the battery with $90\%$, and the jenga-block with $80\%$ success-rate. However, the sizes of the objects are such that they can only be dropped into the mug from the top (subgoal B), however \vanilla is not able to infer these features solely from the training data and hence fails at subgoal B and C. With \ours, the expert language feedback helps establish the correct functional correspondences, leading to an improvement in the performance across all subgoals. 

\subsection{What kind of features maximally improve the sub-goal level performance for observation interventions based methods?}

As shown in \figref{fig:res_id}, all intervention based method demonstrate a $100\%$ task success on all subgoals of the sweep-trash task, in the ID environments. On the object-in-cup task intervention based methods again perform comparably on the $\IDEnv^{marker}$, however on the $\IDEnv^{pen}$ both \policyem and \dino perform worse as compared to \ours on all subgoals.

As shown in \figref{fig:res_bg}, under a novel background, intervention based methods perform comparably on the sweep-trash task. On the object-in-cup task, \policyem performs worse compared to both \dino and \ours, whereas \dino performs comparably with \ours. This can be attributed to the ability of dino features to perform dense correspondence matching, specially across objects in the same semantic class.

\figref{fig:res_ob} shows that under novel objects both \policyem and \dino struggle. Since \policyem relies on the policy embeddings, under `doritos' and `napkin' it sweeps them in either direction. \dino matches the visual features and since napkin closely resembles paper, it is able to correctly sweep napkin as recycling, and fails on other objects.

For the object-in-cup task, because policy embeddings and visual features alone are not enough to match the objects with the ID sample that lead to the desired behavior mode, both \policyem and \dino perform worse as compared to \ours.

\end{document}